\documentclass{article}

\usepackage[T1]{fontenc}
\usepackage{microtype}
\usepackage{graphicx}
\usepackage{subfigure}
\usepackage{booktabs}
\usepackage{hyperref}

\usepackage[accepted]{icml2025}
\makeatletter
\renewcommand{\ICML@appearing}{Preprint.}
\makeatother
\usepackage{amsmath}
\usepackage{amssymb}
\usepackage{mathtools}
\usepackage[capitalize,noabbrev]{cleveref}
\usepackage{xcolor}

\icmltitlerunning{PenGPT}

\usepackage{float}
\usepackage{stfloats}
\newcommand{\FinalHandScore}{0.711}
\newcommand{\FinalSketchScore}{0.588}
\newcommand{\LongerSketchScore}{0.589}

\newcommand{\FinalHandSize}{5.8}
\newcommand{\FinalSketchSize}{49.8}
\newcommand{\FinalHandCost}{0.001}
\newcommand{\FinalSketchCost}{0.003}
\newcommand{\RecurrentRun}{1e-3}

\begin{document}

\twocolumn[
\icmltitle{PenGPT: A Generative Transformer for Handwriting and Sketches}

\begin{icmlauthorlist}
\icmlauthor{Sam Greydanus}{independent_aff}
\end{icmlauthorlist}

\icmlaffiliation{independent_aff}{Independent Researcher}
\icmlcorrespondingauthor{Sam Greydanus}{sam.greydanus@gmail.com}
\icmlkeywords{Transformers, Generative AI, Handwriting, Sketch Generation, Sequence Modeling, Tokenization}

\vskip 0.3in]

\printAffiliationsAndNotice{}

\begin{abstract}
Generative models of text and images have become remarkably capable, but pen strokes, the sequences of pen positions behind handwriting and drawing, have received far less attention. Can one small transformer learn to hold a pen? We introduce PenGPT, a decoder-only model that emits strokes one token at a time, with a simple tokenizer: a walk on a grid, merged by byte-pair encoding. A small 5M-parameter model writes realistic cursive and print in many styles and takes up the style of unseen writers. A larger variant (54M parameters) draws sketches from text. We merge ten public sketch datasets and filter for quality in order to obtain nearly 500k diverse, high-quality line sketches to train on. We ablate each part of the recipe and compare both models against recurrent networks in the style of Graves and of sketch-rnn. Our results show that the gains come mostly from data and training setup: more data, augmentation, and training on several words at once matter most. A BPE vocabulary sixty times larger, from ScribeTokens, does not help. A simple GPT-style architecture trained on large amounts of high quality data gives the best results.
\end{abstract}

\section{Introduction}

\begin{figure}[t!]
    \centering \includegraphics[width=\columnwidth]{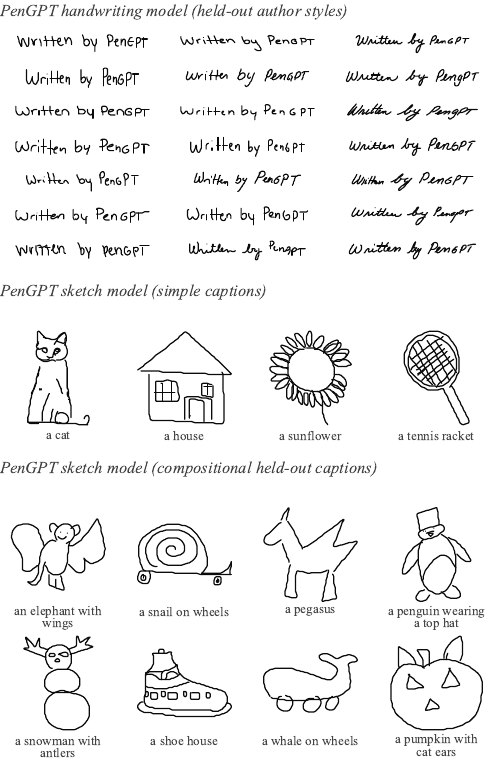}
    \vspace{-1.5\baselineskip}
    \caption{Samples from the two models. Top: one phrase in the hands of 21 writers held out of training; the model reads one to three real words by each writer and continues in their hand. Print to cursive across; neat to messy down, with sampling temperature rising from 0 to 0.6. Middle: drawings of things the model was trained to draw. Bottom: combinations that appear in none of the labels or captions it was trained on. The drawings are the best third of the samples of each prompt, chosen by eye; Figure~\ref{fig:random} has random ones.}
    \label{fig:hero}
\end{figure}

\begin{figure*}[t]
    \includegraphics[width=\textwidth]{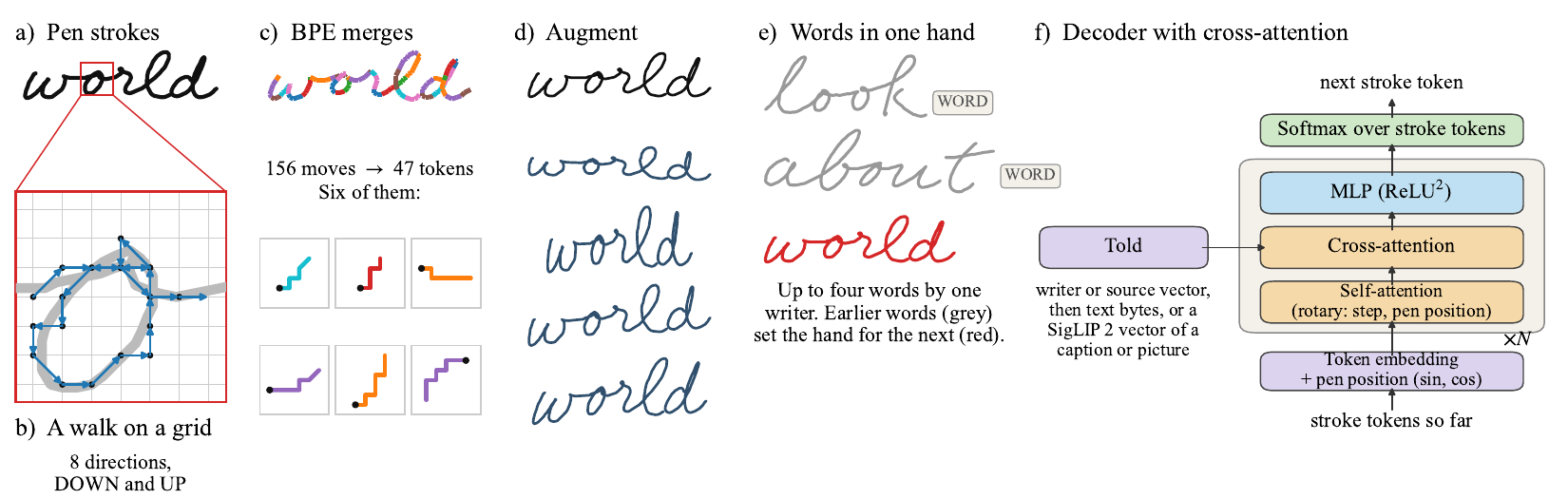}
    \vspace{-1.3\baselineskip}
    \caption{Overview of PenGPT. (a) Data are pen strokes. (b) Points are snapped to a grid and motion becomes a walk in eight directions, with two more symbols for putting the pen down and lifting it. (c) Byte-pair encoding merges common runs of moves, so that one token is a short stroke; six of the tokens in this word are drawn below it, each starting at its dot. (d) Every training sequence is bent by a fresh random affine map. (e) A handwriting sequence holds up to four words by one writer, joined by a \textsc{word} token, which is what lets a few words set the hand at sampling. (f) A decoder-only transformer predicts the next stroke token; what it is told (text bytes or a SigLIP~2 vector, and a learned vector for the writer or source) enters through cross-attention.}
    \label{fig:schema}
\end{figure*}

\begin{figure*}[b]
    \includegraphics[width=\textwidth]{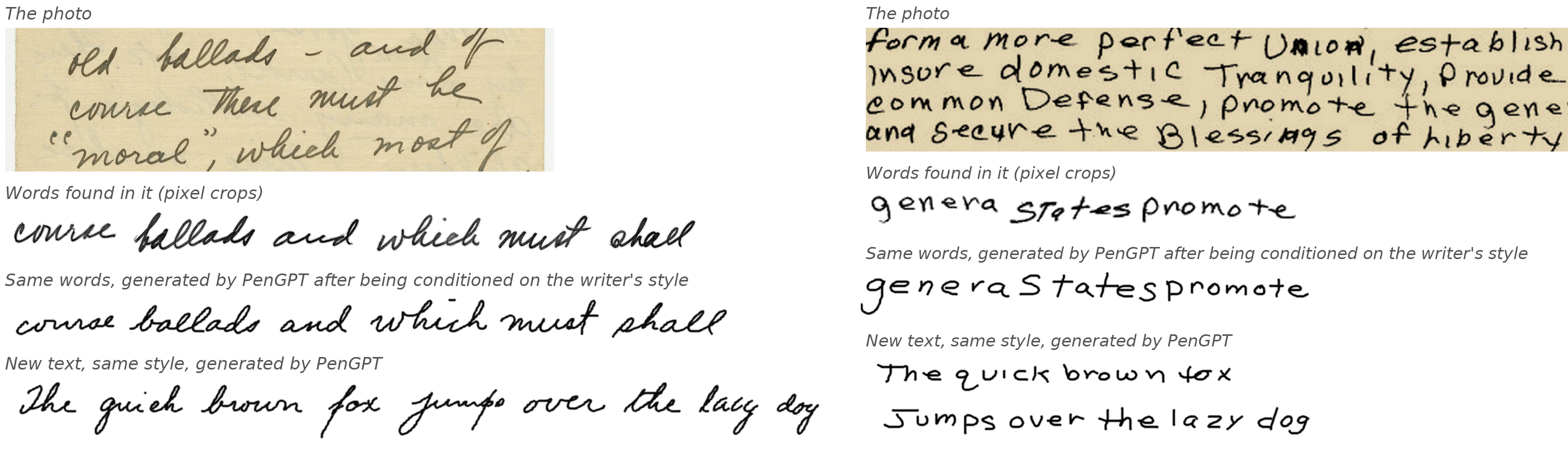}
    \vspace{-1.2\baselineskip}
    \caption{A hand from a photo. Left: a letter by the poet Sara Teasdale. Right: a paragraph printed on a NIST handwriting sample form in 1989, tinted the colour of the letter's paper here so that it reads as a scan. From the top: part of the photo; some of the words found in it; the same words written by the model, after one writer vector is fit to traces of the photo's words with every other weight frozen; and new text in that hand.}
    \label{fig:photo}
\end{figure*}

Neural networks have become remarkably good at generating text, images and audio. Yet one of the oldest ways people record both language and pictures, the trace of a pen, has received far less attention. Handwriting and drawing leave the same record: a sequence of pen positions, with the pen sometimes lifted. That record has structure at every scale, from the direction of a single stroke to the slant a word shares with its neighbours, and all of it must follow what the model is told: the text to write, or a description of the thing to draw.

Earlier work treated this as a problem for special machinery. \citet{graves2014generating} predicted continuous pen offsets with a mixture density network and a self-advancing attention window; \citet{ha2018neural} did the same for sketches inside a variational autoencoder. In the Cursive Transformer \citep{greydanus2025cursive} we argued that a custom \emph{tokenizer} is a better place for that effort than a custom \emph{model}. However, that model wrote in one hand, from 3,500 words, with a vocabulary built by hand. Can one small, standard transformer learn to hold a pen, in many hands and for drawings as well as words?

In this paper we keep the argument and replace nearly everything else (Figure \ref{fig:schema}). Pen points are snapped to a grid, motion becomes a walk in eight directions, and byte-pair encoding \citep{sennrich2016neural} learns the rest of the vocabulary. The data is two orders of magnitude larger, mostly scanned pages traced back into strokes. The model is told who is writing and is trained on several words by one writer at a time, so a few words by a new writer are enough to continue in their hand (Figure \ref{fig:hero}, top). The same code, told a vector from a vision-language model in place of text, draws from a phrase (Figure \ref{fig:hero}).\footnote{Code and weights are at github.com/greydanus/pengpt. We do not release the datasets; Appendix \ref{app:pipeline} describes how they were built.}

Our first contribution is the two models: a 5M-parameter model that writes any text in print or cursive and takes up the hand of an unseen writer, and a 54M-parameter model that draws from typed text. Our main contribution is to ask where their performance comes from. We ablate each part of the recipe, compare against a recurrent network in the style of \citet{graves2014generating}, and introduce \emph{style rank}, a measure of how well a model takes up an unseen hand. The answer is consistent: data, augmentation and the way training sequences are built matter most, while pen-position features and, for handwriting, model size matter little. A recurrent network trained on the same tokens is behind on handwriting by about as much as removing augmentation, and transfers to new writers nearly as well; on drawings, whose sequences are several times longer, it is far behind.

\section{Related Work}

\textbf{Handwriting as strokes.} \citet{graves2014generating} remains the cornerstone: a recurrent network with a mixture density output and an attention window over the text, which can be primed with a real sample to continue in that writer's hand. Later models kept continuous outputs and added explicit machinery for style \citep{aksan2018deepwriting,kotani2020generating,dai2023disentangling,timofeev2023dss}, including for Chinese characters \citep{zhang2018drawing,tang2019fontrnn,tang2021write}; diffusion models \citep{luhman2020diffusion,pan2026diffink} and transformers \citep{jin2025trink} have since replaced the recurrent network. Our model has no style encoder: it emits discrete tokens and takes up a new hand only by continuing a sequence that begins with that writer's words. A larger literature generates handwriting as images (Appendix \ref{app:related}).

\textbf{Tokenizing ink.} Writing a curve as steps between neighbouring grid cells is the chain code of \citet{freeman1961encoding}. The Cursive Transformer \citep{greydanus2025cursive} binned pen offsets in polar coordinates, and digital-ink models give a language model coordinate tokens \citep{fadeeva2024representing,mitrevski2024inksight}. Byte-pair encoding has been applied to music \citep{fradet2023byte} and robot actions \citep{pertsch2025fast}, and quantized motion steps to trajectories \citep{seff2023motionlm,philion2024trajeglish}. Closest to our tokenizer is ScribeTokens \citep{wang2026scribetokens}, which proposes the same recipe and evaluates it on IAM-OnDB. We differ in what is built on the tokens: many writers and transfer to unseen ones, drawings as well as words, two orders of magnitude more data, and ablations. We also run its published BPE in our model and our recipe on its benchmark (Appendix \ref{app:scribetokens}).

\textbf{Sketches.} \citet{ha2018neural} introduced Quick, Draw! and sketch-rnn, a recurrent variational autoencoder trained one class at a time. Later work changed the stroke parameterisation \citep{das2020beziersketch,aksan2020cose} or moved to diffusion \citep{wang2023sketchknitter,das2023chirodiff}; \citet{xu2023deep} survey the field. Transformers over discrete sketch tokens \citep{ribeiro2020sketchformer,tiwari2024sketchgpt,wang2025vq} are trained on one dataset and conditioned on a class at most. Text-driven methods either optimise one drawing at a time against CLIP or a diffusion prior \citep{li2020differentiable,frans2022clipdraw,vinker2022clipasso,jain2023vectorfusion,xing2023diffsketcher}, train on SVG paths \citep{carlier2020deepsvg,wu2023iconshop,tang2024strokenuwa,rodriguez2025starvector,yang2025omnisvg}, or ask a large multimodal model to draw with no training \citep{vinker2025sketchagent,sesay2026handwritingagent}. Ours is a small trained model of pen motion that merges ten datasets and draws any phrase in one sampling pass.

\section{Methods}

\textbf{Tokens.} A word or a drawing is a list of points $(x, y, p)$, where $p$ says whether the pen is down \citep{graves2014generating}. Points are snapped to a grid and the path between them is written as single steps to one of the eight neighbouring cells, with two further symbols, \texttt{DOWN} and \texttt{UP}, for where the pen touches and leaves the page (Figure \ref{fig:schema}b). We then learn byte-pair merges over runs of moves within a stroke, 512 for handwriting and 1,024 for drawings, so that a token is a short stroke with a fixed displacement (Figure \ref{fig:schema}c). A held-out word is 138 moves and 43 tokens on average. The polar tokenizer of \citet{greydanus2025cursive} spends two tokens on every pen point and needs 370 hand-chosen bins; the grid walk has two numbers to choose and is the same for both kinds of data.

\begin{table*}[!t]
\caption{Baselines and ablations. Each row changes one thing about the base recipe; base rows are the mean of three seeds $\pm$ their standard deviation. Score is nats per pen move on held-out data (lower is better) and $\Delta$ is against the base over the sources both runs hold out, so the Quick, Draw! row is compared on Quick, Draw! alone. \textbf{Handwriting:} Unseen is the score for held-out writers after three of their words, Style rank is of 16 (lower is better), and Legible is the share of words, sampled at temperature 1, that TrOCR reads exactly. The ScribeTokens BPE row puts that tokenizer's list of 31,988 merges in place of our 512 and changes nothing else, which makes 60 times the tokens, 3.3 times the parameters and 4.1 times the memory of a training step (Appendix \ref{app:scribetokens}). \textbf{Drawings:} Loss is nats per token; Labels and Captions are how often SigLIP~2 ranks a drawing's own prompt first among 345 typed labels or 300 held-out caption phrases, at guidance 2 ($^\ddagger$guidance 1, where the base scores 54.9\% and 28.8\%). The LSTM drawing baseline is the best of three learning rates (\RecurrentRun). Table \ref{tab:hand_full} has every handwriting measure.}
\label{tab:hand}\label{tab:sketch}
\vskip 0.1in
\begin{center}
\begin{small}
\begin{tabular*}{\textwidth}{@{\extracolsep{\fill}}lccccc}
\toprule
\textbf{Handwriting} & Score & $\Delta$ & Unseen & Style rank & Legible \\
\midrule
Graves-style LSTM, our tokens & 0.745 & $+0.029$ & 0.806 & 2.2 & 49\% \\
\textbf{PenGPT (base)} & \textbf{0.716{\tiny$\,\pm$0.001}} &  & \textbf{0.786{\tiny$\,\pm$0.001}} & \textbf{1.9{\tiny$\,\pm$0.1}} & \textbf{54\%{\tiny$\,\pm$1}} \\
\midrule
no BPE merges ($\sim$2x compute) & 0.706 & $-0.010$ & 0.791 & -- & 55\% \\
ScribeTokens BPE (60x tokens, 3x params, 4x mem) & 0.714 & $-0.001$ & 0.792 & 1.9 & 53\% \\
no pen position & 0.716 & $+0.001$ & 0.786 & 1.9 & 52\% \\
no augmentation & 0.744 & $+0.028$ & 0.804 & 2.0 & 52\% \\
no writer vectors & 0.768 & $+0.053$ & 0.787 & 1.8 & 54\% \\
one word a sequence & 0.724 & $+0.009$ & 1.157 & 6.0 & 52\% \\
\midrule
1.9M param. (63\% smaller) & 0.732 & $+0.016$ & 0.793 & 2.0 & 50\% \\
11.2M param. (114\% larger) & 0.712 & $-0.004$ & 0.786 & 1.9 & 55\% \\
a tenth of the words & 0.765 & $+0.050$ & 0.826 & 3.0 & 47\% \\
without CSAFE & 0.729 & $+0.011$ & 0.813 & 2.2 & 50\% \\
\midrule
real held-out ink &  &  &  &  & 66\% \\
\bottomrule
\toprule
\textbf{Drawings} & Score & $\Delta$ & Loss & Labels & Captions \\
\midrule
LSTM decoder, our tokens & 0.640 & $+0.049$ & 2.940 & 40.3\% & 19.8\% \\
\textbf{PenGPT (base)} & \textbf{0.591{\tiny$\,\pm$0.001}} &  & \textbf{2.712{\tiny$\,\pm$0.004}} & \textbf{75.6\%{\tiny$\,\pm$0.6}} & \textbf{43.3\%{\tiny$\,\pm$0.4}} \\
\midrule
no pen position & 0.599 & $+0.008$ & 2.747 & 73.9\% & 40.3\% \\
added pen position only & 0.593 & $+0.002$ & 2.720 & 75.7\% & 44.7\% \\
rotary pen position only & 0.592 & $+0.001$ & 2.716 & 76.2\% & 40.8\% \\
no augmentation & 0.604 & $+0.013$ & 2.753 & 3.9\% & 6.2\% \\
no stroke-order alterations & 0.596 & $+0.005$ & 2.735 & 75.2\% & 42.2\% \\
no label dropout & 0.591 & $0.000$ & 2.712 & 57.7\%$^\ddagger$ & 32.7\%$^\ddagger$ \\
no ink and stroke hints & 0.590 & $0.000$ & 2.710 & 76.2\% & 43.5\% \\
\midrule
label bytes, no vectors & 0.597 & $+0.006$ & 2.742 & 74.1\% & 0.5\% \\
picture vectors only & 0.580 & $-0.011$ & 2.659 & 6.1\% & 4.3\% \\
picture and label vectors & 0.596 & $+0.005$ & 2.737 & 77.5\% & 9.5\% \\
\midrule
8 layers, 384 wide & 0.599 & $+0.008$ & 2.750 & 71.0\% & 35.8\% \\
6 layers, 256 wide & 0.612 & $+0.022$ & 2.813 & 62.2\% & 30.8\% \\
Quick, Draw! only & 0.522 & $+0.002$ & 2.808 & 73.5\% & 36.7\% \\
\bottomrule

\end{tabular*}
\end{small}
\end{center}
\vskip -0.1in
\end{table*}

\textbf{Data.} The handwriting data is 495,590 words by 1,255 writers from seven sources (Table \ref{tab:data}). Two were recorded as strokes; the rest began as scanned pages: the CEDAR letter \citep{srihari2002individuality}, the CSAFE database \citep{crawford2020csafe}, and letters, diaries and notes from public archives. We hold out 63 writers entirely and 4,744 words by known writers. The drawing data is 477,670 drawings from ten sources, the largest being Quick, Draw! \citep{ha2018neural}, Sketchy \citep{sangkloy2016sketchy} and TU-Berlin \citep{eitz2012humans}.

\textbf{From pages to strokes.} No large collection of English pen strokes by many writers exists, so we make one from scans in four steps (Appendix \ref{app:pipeline}). \emph{Find:} a vision-language model grades a page or two of each archive document for legibility and scan quality; of 4,596 documents graded, about one in fifty is kept. \emph{Cut:} the same model transcribes each page line by line, Kraken \citep{kiessling2019kraken} finds the baselines, and connected blobs of ink are grouped into words so that each line's blobs match its transcript; no blob is ever cut. \emph{Trace:} InkSight \citep{mitrevski2024inksight} turns a word image into strokes; we show it up to four renderings of the image and keep the trace that covers the ink best. \emph{Check:} a trace is kept only if it covers the ink, stays on the ink, draws nothing twice, and reads back as its label. Between a third and a half of the words on the pages survive (31\% of the CEDAR letter, about half of CSAFE), and they lean short and neat: two thirds of the words we keep have four letters or fewer.

\textbf{Choosing and describing drawings.} Crowd-sourced drawings vary enormously in quality, so we score every drawing and keep the best: a vision-language model rated 12,704 drawings, and a linear probe on CLIP embeddings \citep{radford2021learning,schuhmann2022laion} fit to those ratings scores the rest. Of the 50 million drawings in Quick, Draw! we keep the best 0.5\%, and most of the other sources. Each drawing then gets up to 17 captions. A model shown the drawing as it was drawn, in stages, writes a name, a sentence about the whole and a step for each stage; a second model, which sees only that record, rewrites it as fourteen other kinds of caption, from a broad category to a dense description.

\textbf{Sequences and augmentation.} A handwriting training sequence is up to four words by one writer, in random order, with the text of all of them as the prompt (Figure \ref{fig:schema}e). This costs nothing and teaches the model that later words look like earlier ones, which is what lets it hold a hand steady over a page and take up a hand from a few example words (Figure \ref{fig:hero}, top; Figure \ref{fig:transfer} in the appendix), as \citet{graves2014generating} did by priming. Every sequence is bent by a random rotation, shear, stretch and shift (Figure \ref{fig:schema}d); drawings get a stronger version and occasional changes of stroke order. A shift of a fraction of a grid cell changes which cells a stroke crosses, so augmentation changes the tokens and not only the geometry.

\textbf{Model.} The model is a decoder-only transformer \citep{vaswani2017attention} with cross-attention in every block (Figure \ref{fig:schema}f), following current small-GPT practice \citep{karpathy2022nanogpt,jordan2024modded,su2021roformer,so2021primer}. The handwriting model has 8 layers of 192 dimensions (5.2M parameters) and the drawing model 12 layers of 512 (54M). A token says how the pen moves, not where it is, so we add an encoding of the pen's running position to each token embedding; for drawings the same position also turns the rotary encoding in self-attention. Matrices are trained with Muon \citep{jordan2024muon} and everything else with AdamW \citep{loshchilov2019decoupled}; the handwriting model trains in 25 minutes on one A100. We keep the checkpoint with the lowest held-out loss (Table \ref{tab:hypers}). The ablations below use these base recipes; the released models, which drew Figure \ref{fig:hero}, are the same recipes trained for longer (Appendix \ref{app:methods}).

\textbf{What the model is told.} For handwriting the prompt is the bytes of the text. For drawings it is a single SigLIP~2 vector \citep{tschannen2025siglip2}, which during training comes from the drawing's picture, its label or one of its captions; because SigLIP~2 places pictures and text in one space, a typed phrase can stand in at sampling. Both prompts begin with a learned vector for the writer or the source, dropped half the time for handwriting so that the model also learns to write for an unknown writer. Drawings are trained with nothing told 10\% of the time, which allows classifier-free guidance on the logits \citep{ho2022classifier,sanchez2024stay}.

\textbf{Measures.} \emph{Word score} is the negative log-likelihood of a held-out word, in nats per pen move. For writers held out of training we report it after three other words by the same writer. \emph{Style rank} asks whether the model can tell whose hand it is continuing, a likelihood-based form of writer identification \citep{bulacu2007text}: a held-out word is scored after three words by its own writer and after three words by each of 15 other held-out writers from the same source, and we report the rank of the true writer (1 is perfect, 8.5 is chance). \emph{Legibility} is tokenizer-free: each model writes held-out words, sampling at temperature 1, and TrOCR \citep{li2021trocr} reads them back \citep{afonin2023sampling}; real ink read the same way is the mark to reach, not to pass. For drawings we report how often SigLIP~2 ranks a drawing's own prompt first among all prompts, in the spirit of CLIPScore \citep{hessel2021clipscore}.

\section{Results}
\label{results}

\subsection{Handwriting}

\begin{figure*}[t]
    \includegraphics[width=\textwidth]{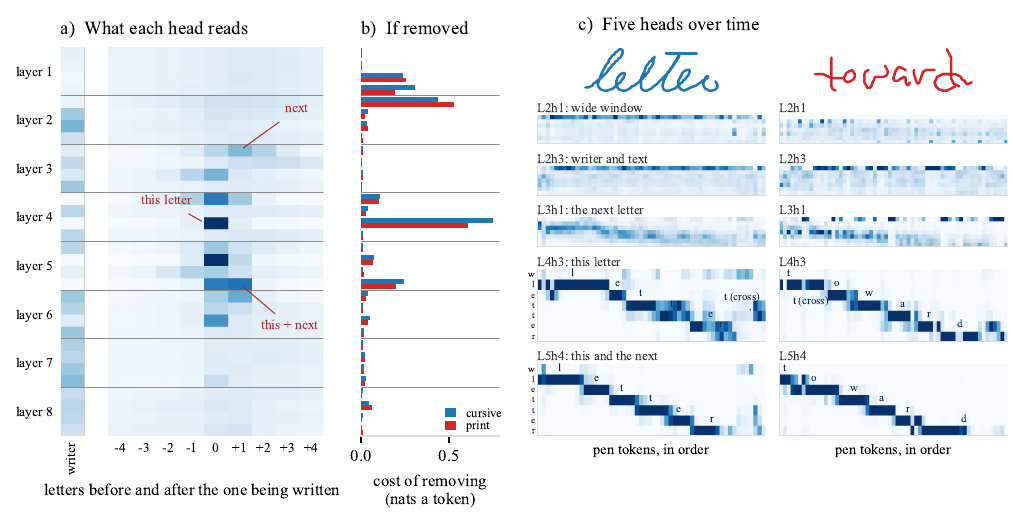}
    \caption{What the cross-attention heads do. (a) Every head of the base handwriting model: its attention on the writer vector and on the letters before and after the one being written, averaged over 1,024 held-out sequences. Positions are measured from the letter that layer 4, head 3 reads. A few heads in layers 3 to 5 each read one position; the rest read a wide window. (b) The cost of silencing each head (its attention set to zero), on the same rows, for the most cursive and the most printed quarter of sequences. (c) Five heads followed over one cursive and one printed held-out word: attention on the writer (w) and on each letter, for every pen token in the order written, on the colour scale of (a). For the two sharpest heads each streak is marked with the letter being drawn; a crossed t is read twice, once for its stem and again for its cross, which in the cursive word comes only after the last letter.}
    \label{fig:heads}
\end{figure*}

The base model scores $0.716$ nats per pen move on known writers, with a standard deviation of $0.001$ over three seeds: the noise to read the other rows of Table \ref{tab:hand} by.

\textbf{Data matters most.} Training on a tenth of the words costs $0.050$ and the model overfits within 6,000 steps; removing the largest source costs $0.011$ on the sources that remain. The model is not short of capacity: doubling its parameters gains $0.004$ and cutting them by nearly two thirds costs $0.016$. Without augmentation the held-out loss rises from halfway through training, and at its best the run is $0.028$ behind.

\textbf{Several words at a time is what makes transfer possible.} A model trained on one word a sequence is only $0.009$ behind on single words, but priming it with three words by the same writer makes it \emph{worse} ($0.816$ to $1.157$), and its style rank worsens from $1.9$ to $6.0$. The writer vector plays no part in transfer: dropping it costs $0.053$ on known writers, the largest effect of any part of the model, and changes nothing on unseen ones. A new hand comes from the words the model is primed with, not from the table of writers (Figure \ref{fig:hero}, top; more in Figure \ref{fig:transfer}).

\textbf{A hand from a photo.} Priming needs strokes, and a photo has none. For a photo of handwriting we box the words, have a vision-language model read them, trace the clearest eight with InkSight, and fit one new writer vector to the traces by 30 steps of gradient descent, with every other weight of the released model frozen (Figure \ref{fig:photo}). So a writer vector can carry a new hand once it is fit to that hand; what does not help with an unseen writer is the table of vectors learned in training. We show this route and do not score it: every measure of transfer in this paper is of priming.

\textbf{The rest matters little.} Removing pen position costs $0.001$; words are short and mostly move left to right. Without byte-pair merges the model is $0.010$ \emph{better} on known writers, at 3.2 times the sequence length and twice the training time, and its sampled words are read as often (55\% against 54\%), so merges are a choice for speed and context. Whose BPE matters as little: ScribeTokens', with 32,000 tokens to our 526, changes the score by $0.001$ and leaves style rank and legibility where they were, with 3.3 times the parameters, four times the memory and 1.7 times as long for a training step. A Graves-style network (three LSTM layers of 400 units with his attention window, 8.5M parameters)\footnote{\label{fn:baselines}The baselines are our own implementations: \citet{graves2014generating} released no weights and we found no code that still runs, and ScribeTokens released its merges, grid and split, which we use word for word, but not its trained model. Its one published number, 72.8\% read exactly, comes from a recogniser trained on the same data and is not on TrOCR's scale (Appendix \ref{app:scribetokens}).} trained on our tokens is $0.029$ behind the transformer on known writers and $0.020$ behind on new ones; it names the true writer first 55\% of the time against 68\%, and its sampled words are read less often (49\% against 54\%).

\textbf{Legibility.} No model's sampled ink is read as often as real held-out ink (66\%); the nearest are the base, the larger model and the model without merges, and the furthest the model trained on a tenth of the words (47\%). We sample because the most likely ink rewards neatness over faithfulness: read that way the base scores 75\%, above real ink, and the model with no writer vectors, which writes every word in one plain hand, scores 87\% (Table \ref{tab:hand_full}). The polar tokenizer of the Cursive Transformer, whose likelihood is not comparable, is in Table \ref{tab:hand_full}.

\textbf{What the heads do.} The cross-attention heads divide the text among themselves by position (Figure \ref{fig:heads}). One head in layer 4 puts 81\% of its attention on the letter being written; another splits evenly between that letter and the next; others read one to three letters ahead; and the heads of the first and last layers read a window of four or five characters, a fifth of it on the writer. Silencing a head, by setting its attention to zero, shows which the model cannot do without. The head that reads the current letter costs the most, $0.6$ to $0.75$ nats a token, and most heads cost less than $0.05$. But sharpness is not the whole story: one wide reader in layer 2 costs $0.4$ to $0.5$ and two in layer 1 cost $0.2$ to $0.3$, while a second sharp head, in layer 5, costs under $0.1$. The heads that read the model's own ink matter more than any of these: silencing one self-attention head in layer 1 or one in layer 2 costs $1.3$ to $2.5$ nats. We looked for heads that serve cursive or print alone and found none: every head reads the same positions in both.

\subsection{Drawings}

\textbf{Guidance.} With no guidance, SigLIP~2 picks a drawing's own prompt out of 345 for 61\% of drawings; at a strength of 2 it does for 77\% (Table \ref{tab:guidance}, four drawings of each label; Table \ref{tab:sketch} draws two and three seeds, and reads 76\%). The gain reverses at 4, and drawings get simpler as guidance rises, from 15 strokes to 9.

\textbf{Ablations.} Augmentation and data matter again, but here so do pen position and size (Table \ref{tab:sketch}). Augmentation matters most to what is drawn: without it the model overfits by step 9,000 of 54,000 and its drawings stop following their prompts, matching 14\% of labels with no guidance and 4\% at guidance 2, which collapses them to four strokes. Pen position matters here, as it did not for handwriting: removing it costs $0.008$ nats and two to three points of prompt agreement, though either form alone recovers nearly all of it. Size matters too: a model with 6 layers of 256 costs $0.022$ and thirteen points on labels. A model trained on Quick, Draw! alone is as good on its own source but worse on caption phrases (37\% against 43\%), which is what merging ten sources buys. The rest is within noise: hints of how much ink to use, changes of stroke order, and label dropout, which costs nothing in likelihood and is what allows guidance. The model matters more here than for handwriting. A three-layer LSTM decoder on the same tokens (59M parameters), told the SigLIP~2 vector at every step as in sketch-rnn \citep{ha2018neural}, is $0.049$ behind at the best of three learning rates, and its drawings match their labels 40\% of the time against 76\%, and caption phrases 20\% against 43\%.

\begin{figure*}[t]
    \includegraphics[width=\textwidth]{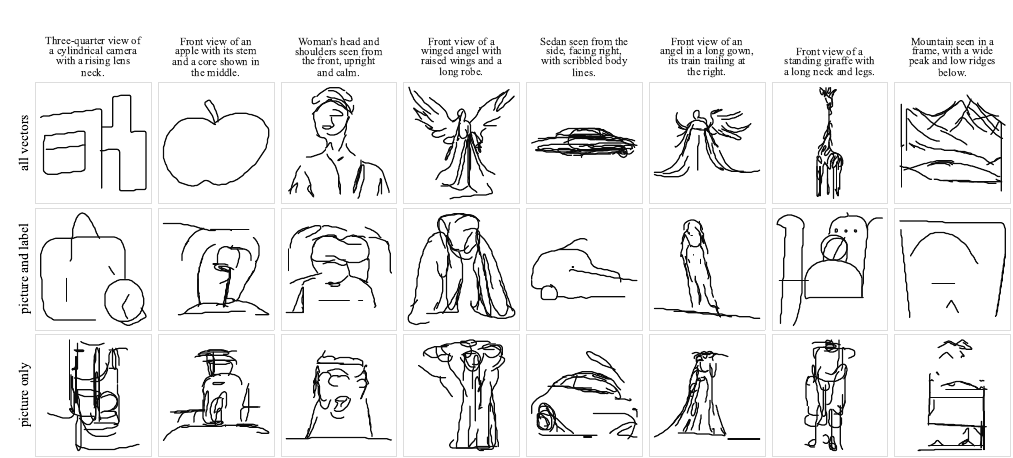}
    \vspace{-1.2\baselineskip}
    \caption{Are the captions needed? Eight held-out caption phrases, each drawn once by three models that differ in what they were told in training, at guidance 2 and temperature $0.6$, with nothing picked or redrawn. Only the model trained on captions follows the phrase. Figure \ref{fig:captions} has sixteen.}
    \label{fig:captions_top}
\end{figure*}

\textbf{Are the captions needed?} Captions are the costly part of the data: they are written by a vision-language model and multiply its size. Since SigLIP~2 places pictures and text in one space, a cheaper model might be trained on picture vectors alone and told a text vector when it draws. We trained three models that differ in one flag (all vectors, picture and label, picture only) and had each draw typed prompts.

The captions are needed for free text and for nothing else. With all vectors a drawing's own prompt is ranked first for 75\% of labels and 43\% of caption phrases. The model told pictures and labels matches it on labels (78\%) and fails on phrases (10\%). The model told pictures alone fails on both (6\% and 4\%), drawing long scribbles of 84 strokes where the others draw 15.

Likelihood shows none of this: on held-out pen moves the picture-only model is the \emph{best} of the three ($0.580$ against $0.591$), because the picture is the most informative thing a model can be told. The failure is one of geometry. SigLIP~2 vectors of sketches sit in a narrow cone (the cosine between two pictures is $0.73$) nearly at right angles to text (a picture's cosine to its own label is $0.11$), so a model that has only seen the cone is far outside its training range when told text (Figure \ref{fig:captions_top}; Appendix, Tables \ref{tab:typed} and \ref{tab:geometry}). We did not test which of the 17 kinds of caption matter; one or two may do.

\section{Discussion}
\label{discussion}

\textbf{Custom tokenizers instead of custom models.} The argument of \citet{greydanus2025cursive} was that recasting a niche modality as tokens lets one train a standard model on it. The grid walk makes the same argument with less design: ten symbols and a merge count, against several hundred hand-placed bins. That it serves handwriting and drawings unchanged is some evidence that the tokenizer is the right place for the modality-specific work.

\textbf{Where the performance comes from.} The ablations point away from the model, as similar studies have in vision and language \citep{sutton2019bitter,wightman2021resnet,steiner2022how}. Pen position and merges move the handwriting score by $0.01$ or less, and doubling the model by less than that; data, augmentation and the way sequences are built move it by more. Replacing the transformer with a recurrent network costs $0.03$, as much as augmentation, yet the recurrent model still takes up a new hand (style rank $2.2$ against $1.9$): the tokens and the training sequences carry over to a different model. Drawings are the exception. Their sequences are several times longer, and there both size and the choice of model matter: a recurrent decoder draws what it is asked about half as often. The single most useful decision was to train on several words by one writer at a time, which costs nothing and is the whole of the model's ability to take up a new hand.

\textbf{Limitations.} The drawing model is told one pooled vector, which is close to a bag of words: captions that differ only in ``facing left'' and ``facing right'' have a cosine similarity of $0.999$. The model composes where a blend of two ideas is enough (a teapot with a face) and fails where words must be bound to each other (who wears what). Our drawings are scored by the same family of model that conditions them, so prompt agreement may flatter a model that has learned what SigLIP~2 likes to see. Most of our handwriting was traced from scans by another model, so its stroke order is an inference, not a record. All baselines are our own implementations (footnote \ref{fn:baselines}), and the handwriting baseline was trained at one learning rate. The one comparison on public data, with ScribeTokens on its split of DeepWriting, is in Appendix \ref{app:scribetokens}, and it is read by TrOCR and not by the recogniser that paper used. Sampled at full temperature our handwriting is read less often than real ink (54\% of words against 66\%), and the words we train on lean short and neat. Each ablation is a single run, read against the noise of three base seeds.

\textbf{Closing thoughts.} Much of our earlier work was built around a clever idea, a habit that \citet{sutton2019bitter} warns against. This project had none. The work we compare against put its effort into the model \citep{graves2014generating,ha2018neural} or into the tokenizer \citep{greydanus2025cursive,wang2026scribetokens}; we put ours into the data, and the ablations say that was the right choice. Most of our time went into gathering as many strokes as we could and checking their quality. Most of the rest went into making training fast on a node of eight GPUs, where the handwriting model trains in 8 minutes and the drawing model in about 35; the Cursive Transformer, by contrast, took upwards of ten hours on one GPU and learned a single hand. The tokenizer is simpler and less elegant than the one it replaces, and it serves both kinds of data unchanged. The bitter lesson sidelines many appealing ideas, but it leaves a good deal to study. A model grown from data can be examined as an organism is: what it can do, how it scales and how it works inside.

\section*{Impact Statement}

The goal of our paper is to advance the field of machine learning. One consequence deserves a mention: a model that takes up a person's hand from a few words could be used to imitate their writing, a risk we consider modest beside that of generative models of faces, voices and video.

\bibliography{pengpt}
\bibliographystyle{icml2025}

\newpage
\appendix
\onecolumn
\raggedbottom
\section{Method details}
\label{appendix_a}
\label{app:methods}

\textbf{Tokens.} Handwriting is set on a common frame: the baseline at $y=0$, an x-height of $0.123$ and points about $0.01$ apart; a drawing is scaled so that its longer side is 1. The grid is $0.02$ for handwriting, about a sixth of the x-height, and $0.0125$ for drawings, 80 cells across. With four special tokens (\texttt{PAD}, \texttt{END}, \texttt{WORD}, \texttt{BOS}) the handwriting vocabulary has 526 entries. Decoding replays the walk and smooths each stroke with a three-point average. The polar tokenizer we compare against uses 220 angle bins and 150 radius bins.

\textbf{Data.} One writer in twenty of each handwriting source with at least twenty writers is held out entirely, and one drawing in fifty. The other drawing sources are drawings of THINGS \citep{mukherjee2026drawings}, SEVA \citep{mukherjee2023seva}, Photo-Sketching \citep{li2019photo}, DifferSketching \citep{xiao2022differsketching}, Creative Sketch \citep{ge2021creative}, ControlSketch \citep{arar2025swiftsketch} and line icons from the open icon sets collected by Iconify.

\textbf{Augmentation.} Handwriting sequences get a rotation of up to $2^\circ$, a horizontal shear of up to $0.10$, a stretch of up to 10\% in each axis and a shift of less than one grid cell, shared by the words of a sequence. Drawings get a rotation of up to $6^\circ$, a perspective tilt, a smooth warp and a change of size and position and, with probability $0.15$ each, three changes of drawing order: two neighbouring strokes swapped, a stroke reversed, and a closed stroke started elsewhere on its loop.

\textbf{Model and training.} Blocks use RMS normalization before each sublayer and on queries and keys, ReLU$^2$ MLPs, no biases, and logits capped at $\pm15$ by a tanh. The pen's running position is encoded with sines and cosines at eight scales. The SigLIP~2 vector is projected to four prompt tokens, and drawings are sometimes told how much ink and how many strokes to use. Training uses a warmup of 200 steps and a linear decay over the last 80\%; the handwriting model trains for 60,000 steps of 128 sequences and the drawing model for 144,000 steps of 65,536 tokens.

\textbf{Sampling.} Long text is written two words at a time, each call continuing from the last word written. Handwriting figures use the most likely token at each step unless noted; drawings are sampled at temperature $0.6$ with guidance 2 unless noted. The released models are the base recipes trained for longer: the handwriting model is 208 wide (6.1M parameters) and scores \FinalHandScore{} nats per pen move on known writers, and the drawing model is trained for 144,000 steps and scores \FinalSketchScore{}; a third more steps did not improve it (\LongerSketchScore). Their weights are rounded to 8 bits, which costs \FinalHandCost{} and \FinalSketchCost{} nats per token, giving files of \FinalHandSize{}~MB and \FinalSketchSize{}~MB that run from a single numpy file.

\textbf{Measures.} Word score divides by base pen moves and not tokens, which makes it comparable across vocabularies built on the same grid. Style-rank distractors come from the same source as the true writer so that the model cannot succeed by telling sources apart. Legibility is exact match over letters and digits, in either case, on 200 held-out words from each source. In the caption study, adding the mean difference between picture and text vectors to a typed vector recovers part of the picture-only model's loss (22.3\% of labels and 17.3\% of phrases); the phrases are held-out captions of a kind the full model trained on.

\section{The data pipeline}
\label{app:pipeline}

\subsection{Handwriting}

\textbf{Finding pages.} CEDAR and CSAFE are pages written for research, each writer copying set passages. For the rest we search the Internet Archive, the Library of Congress and public repositories of handwritten study notes. A vision-language model (Claude) is shown one or two pages of each document, says whether the page is handwritten running text in English, and grades from 1 to 10 how easily the hand can be read and how clean the scan is; a page that passes is also asked whether it is set in a handwriting-style font. Of 4,596 documents graded (2,760, 1,108 and 728 from the three places), about one in fifty passes at legibility 7 and scan 6. Nine documents graded 5 or 6 were looked over and judged too messy to use. We take up to 20 pages a writer.

\textbf{Cutting pages into words.} The model transcribes each page line by line, keeping the writer's spelling and leaving out words that are crossed out, and grades its neatness; messy pages are dropped. Kraken finds the baseline of every line. Ink is separated from paper by a threshold after the background is flattened, and each connected blob of ink is given to the line whose band it lies in; a blob shared by two lines marks its word as tangled, and the word is dropped. Lines of blobs are matched to lines of the transcript by their width and the number of words each seems to hold. Within a line a dynamic program groups the blobs into the transcript's words, so that each word is about as wide as its letters ask and the breaks fall at the widest gaps. A blob is never cut, so two words joined by a stroke are lost together. A two-page spread is split at its gutter first: on one letterbook this took the share of word images with the right label from 4\% to 68\%.

\textbf{Tracing.} InkSight reads a word image and writes pen strokes, along with its own reading of the word. We first show it a monoline rendering, the skeleton of the ink thickened to the median stroke width, and give it no text. If that trace fails the checks below it is shown three more: the raw image, and an image with the paper cleaned to white and the monoline rendering again, these two with the label given. Of the traces that pass we keep the one that covers the most ink while staying on it, at 2.1 traces a word on average. Two strokes are joined when the gap between them is short and lies on ink, since InkSight often lifts the pen where the writer did not.

\textbf{Checks.} A trace is drawn at the width of the ink and kept only if all of these hold: it covers at least 80\% of the ink; at least 90\% of its points lie on ink; no connected piece of ink has more than 35\% of its trace drawn twice; the path is no longer than 1.6 times the ink would need; every dot and punctuation mark the label calls for is covered; and InkSight's reading is within a third of the label's letters, and not the label with its first or last letter missing. Coverage may fall to 70\% when the reading is nearly exact, which saves good traces of thick ink. A dot or mark the trace skipped is added as a short stroke across its ink. A word that passes every check but the reading is drawn from its strokes and read again by the vision-language model; when that reading agrees with InkSight's the word takes it as its label, which was right for 95.5\% of 300 words whose labels were known.

\textbf{Yield.} Table \ref{tab:yield} follows one source through. Longer words are lost more often: in a sample of CEDAR, 79 to 83\% of words of two or three letters were accepted, 50 to 58\% of seven or eight, and 14\% of nine or more. Of 64 accepted words drawn at random over their images, none had a wrong label or a trace that missed the word, and three carried a stray mark from a neighbouring word.

\textbf{One frame.} Each writer's words are scaled together so that the x-height is $0.123$ and the baseline is at $y=0$, every stroke is resampled to points $0.01$ apart, and exact duplicates are dropped. The two sources recorded as strokes skip tracing: tablet notes are grouped into words and lines by the gaps between strokes and read by the vision-language model, and only lines of plain dictionary words are kept.

\subsection{Drawings}

\textbf{A quality score.} A vision-language model (Claude) rated 12,704 drawings from all sources, shown eight at a time with their labels, from 1 to 10 for how recognisable, complete and well proportioned they are, with instructions to judge how well a drawing is drawn and not how much is drawn. Each drawing is rated twice, once among drawings of its own source and label and once among drawings of that label from every source; the two ratings correlate at about $0.8$. A linear probe on CLIP ViT-L/14 embeddings of the rendered drawings, in the manner of the LAION aesthetic predictor, is fit by ridge regression on better-against-worse pairs inside one source and label, so that the source and the subject cancel and what is left is how well the thing is drawn. On labels held out of the fit it ranks the drawings of a group as the judge does at a rank correlation of $0.60$; ink length and stroke count alone reach $0.13$, and on a source left out of the fit it reaches $0.51$ to $0.57$.

\textbf{Quick, Draw!} holds 50 million drawings, many of them abandoned or unrecognisable. A drawing must be one the game recognised, with ink length between the 50th and 98th percentile of its class and a stroke count between the 25th and 99th; a small classifier trained on Quick, Draw! itself must then name its class with at least 50\% confidence. What remains is ranked by the quality score plus half the log of that confidence, and each of the 345 classes keeps its best, with no class more than twice the size of another. Rated blind by the judge beside random drawings of the same class, the kept drawings average $6.2$ against $3.9$ and come out ahead in 99\% of 160 classes. From the other sources we keep the best of each label, all of it for most sources, 90\% for THINGS and 25\% for SEVA.

\textbf{Captions.} A drawing is cut into one to five stages where the pen jumps, and shown to a vision-language model one stage at a time with earlier strokes in grey, along with the source's label as a hint. The model writes a \emph{name} of up to four words, a \emph{frame} (one sentence giving the subject, the viewpoint, the pose and the manner of drawing) and one \emph{step} for each stage. A second model sees only that record, not the drawing, and writes fourteen more captions from it, among them a \emph{phrase} of three to six words, a \emph{short} sentence, a longer \emph{sentence} that leads with the manner or the view, a \emph{dense} run of clipped phrases, a request as someone might type it, and a list of tags. A caption is thrown away if it runs over its word limit, gives a count the record does not give, or is too long for SigLIP~2 to read whole. The second model is told to say only what the record states, so that a caption adds no part the first model did not see.

\begin{table}[H]
\caption{Yield. Left: the CEDAR letter, one page from each of 1,568 writers; the words kept all come from the 697 pages that were cut. Right: Quick, Draw!}
\label{tab:yield}
\vskip 0.1in
\begin{center}
\begin{small}
\begin{tabular}{lr@{\hskip 3em}lr}
\toprule
Handwriting (CEDAR letter) & Words & Drawings (Quick, Draw!) & Drawings \\
\midrule
On the pages & about 245,000 & In the dataset & 50.4M \\
Cut as word images, from 697 pages & 99,962 & Recognised, ink and strokes in range & 20.0M \\
Traced and accepted & 76,212 & Named by the classifier & 14.1M \\
 & & Kept & 252,148 \\
\bottomrule
\end{tabular}
\end{small}
\end{center}
\end{table}

\section{Further related work}
\label{app:related}

\textbf{Handwriting as images.} A larger literature generates images of handwriting with GANs \citep{alonso2019adversarial,fogel2020scrabblegan,kang2020ganwriting}, transformers \citep{bhunia2021handwriting,wang2022approach,pippi2023handwritten} and diffusion \citep{nikolaidou2023wordstylist,dai2024one,nikolaidou2024diffusionpen}; \citet{pippi2025zero} imitate an unseen style by continuing a prompt, the image-domain counterpart of what we do. Strokes, unlike pixels, can be restyled, scaled and plotted, and are a far smaller thing to model.

\section{Data and hyperparameters}

\begin{table}[H]
\caption{Training data. Handwriting is CSAFE \citep{crawford2020csafe}, the CEDAR letter \citep{srihari2002individuality}, letters and diaries from the Internet Archive and the Library of Congress, study notes from public GitHub repositories, as scanned pages and as tablet strokes, and the trackpad words of \citet{greydanus2025cursive}. Drawings are Quick, Draw! \citep{ha2018neural}, Sketchy \citep{sangkloy2016sketchy}, ControlSketch \citep{arar2025swiftsketch}, line icons from the open sets collected by Iconify, THINGS \citep{mukherjee2026drawings}, SEVA \citep{mukherjee2023seva}, TU-Berlin \citep{eitz2012humans}, Creative Sketch \citep{ge2021creative}, Photo-Sketching \citep{li2019photo} and DifferSketching \citep{xiao2022differsketching}. Kept is the share of a source's drawings that the quality score keeps, the best of each label first.}
\label{tab:data}
\vskip 0.1in
\begin{center}
\begin{small}
\begin{tabular}{lrr@{\hskip 3em}lrr}
\toprule
Handwriting source & Words & Writers & Drawing source & Drawings & Kept \\
\midrule
CSAFE & 344,438 & 449 & Quick, Draw! & 252,148 & 0.5\% \\
CEDAR letter & 76,212 & 697 & Sketchy & 64,569 & all \\
Internet Archive letters and diaries & 39,435 & 43 & ControlSketch & 34,989 & all \\
Library of Congress manuscripts & 18,353 & 43 & Line icons & 33,288 & all \\
GitHub study notes (scanned) & 11,568 & 13 & THINGS & 25,946 & 90\% \\
Trackpad words & 3,500 & 1 & SEVA & 21,277 & 25\% \\
GitHub tablet notes (strokes) & 2,084 & 9 & TU-Berlin & 20,000 & all \\
 & & & Creative Sketch & 17,164 & all \\
 & & & Photo-Sketching & 4,850 & all \\
 & & & DifferSketching & 3,439 & all \\
\midrule
Total & 495,590 & 1,255 & Total & 477,670 & \\
\bottomrule
\end{tabular}
\end{small}
\end{center}
\end{table}

\begin{table}[H]
\caption{Model and training hyperparameters.}
\label{tab:hypers}
\vskip 0.1in
\begin{center}
\begin{small}
\begin{tabular}{lcc}
\toprule
 & Handwriting & Drawings \\
\midrule
Layers, width, heads & 8, 192, 4 & 12, 512, 8 \\
Parameters & 5.2M & 54M \\
Grid & 0.02 & 0.0125 \\
BPE merges & 512 & 1,024 \\
Block size (tokens) & 256 & 768 \\
Words a sequence & up to 4 & 1 drawing \\
Prompt & text bytes (up to 50) & one SigLIP 2 vector, as 4 tokens \\
Writer or source dropout & 0.5 & 0.1 \\
Prompt dropout (for guidance) & none & 0.1 \\
Pen position & added, 8 bands & added and rotary \\
Batch & 128 sequences & 65,536 tokens \\
Steps & 60,000 & 144,000 \\
Muon learning rate (block matrices) & 0.04 & 0.04 \\
AdamW learning rate (embeddings, rest) & 0.06, 0.006 & 0.006, 0.006 \\
Weight decay & $10^{-4}$ & $10^{-4}$ \\
Warmup, decay & 200 steps, linear over last 80\% & same \\
Precision & 16-bit forward pass & same \\
Held out & 1 writer in 20; 4,744 words & 1 drawing in 50 \\
\bottomrule
\end{tabular}
\end{small}
\end{center}
\end{table}

\section{Handwriting: every measure}

\begin{table}[H]
\caption{Handwriting: every measure for every run. Each row changes one thing about the base recipe and trains for the same 60,000 steps; the Graves-style network trains on the same data as 30,000 larger steps on two GPUs. Word score is nats per pen move on held-out words by known writers (lower is better) and $\Delta$ is against the base recipe over the sources both hold out. The base row is the mean of three seeds, $\pm$ their standard deviation. Unseen writers were held out of training entirely. Style rank is of 16 (lower is better), after three priming words and after one; with no merges, three priming words rarely fit the model's window. Legibility is the share of words TrOCR reads exactly, written as the most likely ink and sampled at temperature 1; Table \ref{tab:hand} reports the second. $^\dagger$The polar tokenizer of the Cursive Transformer quantizes ink differently, and the ScribeTokens grid is finer than ours, so these word scores are not comparable with the others. The polar model's most likely ink often never ends a word; sampled, it is read as often as the base model's.}
\label{tab:hand_full}
\vskip 0.1in
\begin{center}
\begin{small}
\setlength{\tabcolsep}{3pt}
\begin{tabular}{lcccccccc}
\toprule
 & \multicolumn{2}{c}{Known writers} & \multicolumn{2}{c}{Unseen writers} & \multicolumn{2}{c}{Style rank} & \multicolumn{2}{c}{Legibility} \\
\cmidrule(lr){2-3}\cmidrule(lr){4-5}\cmidrule(lr){6-7}\cmidrule(lr){8-9}
 & Score & $\Delta$ & Alone & After 3 & 3 words & 1 word & Most likely & Sampled \\
\midrule
Graves-style LSTM, our tokens & 0.745 & $+0.029$ & 0.828 & 0.806 & 2.25 & 2.69 & 75.0\% & 48.6\% \\
polar tokens (Cursive Transformer) & 2.091$^\dagger$ & -- & 2.123 & 1.935 & 2.46 & 3.70 & 30.0\% & 56.9\% \\
ScribeTokens BPE and grid & 0.570$^\dagger$ & -- & 0.613$^\dagger$ & 0.592$^\dagger$ & 2.11 & 2.80 & 69.5\% & 55.0\% \\
PenGPT (base) & 0.716{\tiny$\,\pm$0.001} &  & 0.813{\tiny$\,\pm$0.001} & 0.786{\tiny$\,\pm$0.001} & 1.87{\tiny$\,\pm$0.07} & 2.53{\tiny$\,\pm$0.06} & 74.8\%{\tiny$\,\pm$0.5} & 54.5\%{\tiny$\,\pm$1.0} \\
\midrule
no BPE merges ($\sim$2x compute) & 0.706 & $-0.010$ & 0.821 & 0.791 & -- & 2.39 & 62.2\% & 55.2\% \\
ScribeTokens BPE (60x tokens) & 0.714 & $-0.001$ & 0.820 & 0.792 & 1.95 & 2.45 & 65.8\% & 53.4\% \\
no pen position & 0.716 & $+0.001$ & 0.814 & 0.786 & 1.85 & 2.37 & 75.4\% & 52.2\% \\
no augmentation & 0.744 & $+0.028$ & 0.827 & 0.804 & 2.04 & 2.80 & 75.3\% & 52.4\% \\
no writer vectors & 0.768 & $+0.053$ & 0.814 & 0.787 & 1.80 & 2.45 & 87.1\% & 53.6\% \\
one word a sequence & 0.724 & $+0.009$ & 0.816 & 1.157 & 5.97 & 6.08 & 75.1\% & 52.5\% \\
\midrule
1.9M param. (63\% smaller) & 0.732 & $+0.016$ & 0.818 & 0.793 & 2.02 & 2.49 & 75.9\% & 50.5\% \\
11.2M param. (114\% larger) & 0.712 & $-0.004$ & 0.814 & 0.786 & 1.91 & 2.67 & 73.6\% & 55.0\% \\
a tenth of the words & 0.765 & $+0.050$ & 0.848 & 0.826 & 2.95 & 3.57 & 72.4\% & 47.2\% \\
without CSAFE & 0.729 & $+0.011$ & 0.836 & 0.813 & 2.16 & 2.61 & 70.3\% & 50.3\% \\
\midrule
real held-out ink &  &  &  &  &  &  & 66.1\% & 66.1\% \\
\bottomrule

\end{tabular}
\end{small}
\end{center}
\vskip -0.1in
\end{table}

\section{Comparison with ScribeTokens}
\label{app:scribetokens}

ScribeTokens \citep{wang2026scribetokens} builds its tokens as we do, from eight directions, pen down and pen up, with byte-pair merges that never cross a pen token. It differs in three settings: a vocabulary of 32,000 where ours has 526, a grid about three and a half times finer, and no scaling of the ink to a common size. To see what those settings are worth we ran our recipe on its benchmark with its BPE.

\textbf{Data.} The benchmark is the part of DeepWriting \citep{aksan2018deepwriting} that comes from IAM-OnDB \citep{liwicki2005iam}: 46,140 words by 200 writers, split at random into 36,912 for training, 4,614 for validation and 4,614 for testing, so the test writers are seen in training. ScribeTokens publishes the split and we use it word for word. The ink goes through our own standardisation (Appendix \ref{app:pipeline}).

\textbf{Runs.} Each run is the base handwriting recipe trained on the training words alone for 30,000 steps, keeping the checkpoint with the lowest loss on the validation words; without dropout all of them reach it by step 6,000 and overfit afterwards. For ScribeTokens BPE we load the list of 31,988 merges its authors publish into our tokenizer. Their grid is a step of 8 in the units of the recording, where the median x-height is 169; in our frame that is a grid of $0.0058$. At that grid and with their BPE a word is about 56 tokens where ours is 43, and the larger vocabulary adds twelve million parameters, in the embedding and the output layer, to a model of five million. The two runs on the finer grid differ in one more thing: ours has a window of 384 tokens and the one with the larger vocabulary, which would not fit in memory at that length, 256.

\textbf{Measures.} We report two numbers on the test words. The first is nats per pen move, which compares runs that share a grid whatever their vocabulary. The second needs no tokenizer: each model writes the test words, sampling at temperature 1 as ScribeTokens does, and TrOCR reads them back; we give the share read exactly, beside the share for the real ink.

\begin{table}[H]
\caption{PenGPT and ScribeTokens BPE on the ScribeTokens split of DeepWriting. Nats per pen move compare only within a grid. Read exactly is the share of 4,614 test words, sampled at temperature 1, that TrOCR reads as their label. The first row is the mean of two seeds, which read 56.5\% and 61.3\%: differences of a few points between single runs are within that spread. The two rows of three seeds are the comparison of systems. ScribeTokens reports 72.8\% for its own model of 34M parameters, but read by a recogniser its authors trained on this data, which reads the real test ink at 83.4\% where TrOCR reads it at 77.7\%; the two shares are not on one scale, so it has no row.}
\label{tab:scribetokens}
\vskip 0.1in
\begin{center}
\begin{small}
\begin{tabular}{llccc}
\toprule
BPE (tokens) & Grid & Parameters & Nats a pen move & Read exactly \\
\midrule
PenGPT (526) & $0.02$ & 5.2M & 0.800 & 58.9\% \\
ScribeTokens (32,002) & $0.02$ & 17M & 0.803 & 61.1\% \\
\midrule
PenGPT (526) & $0.0058$ & 5.2M & 0.621 & 64.5\% \\
ScribeTokens (32,002) & $0.0058$ & 17M & 0.628 & 61.1\% \\
\midrule
PenGPT (526), with dropout $0.2$ (three seeds) & $0.02$ & 5.2M & 0.775 & 69.0\%{\tiny$\,\pm$0.5} \\
ScribeTokens-style model (three seeds) & $0.0058$ & 53M & 0.624 & 67.0\%{\tiny$\,\pm$0.7} \\
\midrule
PenGPT (526), trained on our seven sources only & $0.02$ & 5.2M & 0.906 & 56.6\% \\
real test ink & & & & 77.7\% \\
\bottomrule
\end{tabular}
\end{small}
\end{center}
\end{table}

\textbf{Findings.} Whose BPE it is makes no difference we can measure (Table \ref{tab:scribetokens}). At either grid our 526 tokens fit the test ink slightly better than 32,000, by $0.003$ and $0.008$ nats a pen move, and the share of sampled words read exactly does not separate them beyond the spread between two seeds. The larger vocabulary makes a training step 1.7 times as long. A finer grid may make sampled ink a little easier to read, by about three points. What this small set needs is regularisation: dropout of $0.2$, which ScribeTokens uses and our recipe does not, moves the best checkpoint from step 5,000 to the end of training, improves the fit by $0.025$ and brings the share read exactly to 69.0\% over three seeds, which read 68.8\%, 69.5\% and 68.6\%. Stronger augmentation added nothing to that, and a model of twice the size was worse. Against that we set a model of ScribeTokens' own shape in our code: 12 layers of 384 with dropout of $0.2$, its BPE and its grid, 53M parameters where the paper's has 34M because ours cross-attends in every block and does not tie its embeddings. It reaches its best checkpoint by step 6,500 and overfits afterwards even with dropout, and its sampled words are read exactly 67.0\% of the time over three seeds against 69.0\% for ours, two points apart where the seeds of either differ by about one. The model of Table \ref{tab:hand}, which never saw IAM-OnDB, writes these words nearly as readably as the models trained on them.

\textbf{On our own data.} The same two changes are rows of Table \ref{tab:hand}, trained on our seven sources by the base recipe for its 60,000 steps. With ScribeTokens BPE in place of ours the model scores $0.714$ against $0.716$, takes up an unseen hand as well (style rank $1.9$ for both, though only 115 of its test words left room for three priming words where the base had 304) and is read as often when sampled (53\% against 54\%). The larger vocabulary takes the parameters from 5.2M to 17M and, with each stage of a training step timed apart on one idle A100, the step from 35 to 59 milliseconds and its memory from 3.1 to 12.8~GB. Table \ref{tab:hand_full} also has ScribeTokens BPE on ScribeTokens' grid, with a window of 320 tokens to leave room for priming words. Its scores are on the finer grid and do not compare. Its sampled words are read as often as the base's (55\% against 54\%) and its most likely ink less often (70\% against 75\%). It takes up an unseen hand a little less well, with a style rank of $2.8$ against $2.5$ after one priming word; after three the rank rests on only 48 words, since on that grid three priming words seldom fit even the wider window.

\section{Guidance}

\begin{table}[H]
\caption{Guidance at sampling. Guidance 2 gets nearly all of the gain. 345 Quick, Draw! labels, four drawings of each, temperature $0.6$. A drawing counts when SigLIP~2 ranks its own prompt first, or in the first five, among the 345.}
\label{tab:guidance}
\vskip 0.1in
\begin{center}
\begin{small}
\setlength{\tabcolsep}{4pt}
\begin{tabular}{lcccc}
\toprule
Guidance & Cosine & First & Top 5 & Strokes \\
\midrule
1 & 0.1216 & 60.7\% & 81.0\% & 14.6 \\
1.5 & 0.1299 & 74.3\% & 90.1\% & 11.7 \\
2 & 0.1321 & 76.7\% & 91.7\% & 11.1 \\
3 & 0.1328 & 77.9\% & 91.8\% & 9.3 \\
4 & 0.1305 & 76.0\% & 89.6\% & 9.5 \\
\bottomrule

\end{tabular}
\end{small}
\end{center}
\vskip -0.1in
\end{table}

\section{The caption study}

\begin{table}[H]
\caption{Typed prompts, the caption study in full. Labels are enough for labels; only captions make phrases work. Two drawings of each prompt at temperature 0.6, scored by SigLIP~2 against every prompt of the set. First is how often a drawing's own prompt is the best match; chance is 0.3\%. Shifted adds the mean picture vector less the mean text vector to each typed vector.}
\label{tab:typed}
\vskip 0.1in
\begin{center}
\begin{small}
\begin{tabular}{llcccccccc}
\toprule
 & & \multicolumn{4}{c}{345 labels} & \multicolumn{4}{c}{300 held-out caption phrases} \\
Told in training & Guidance & Cosine & First & First 5 & Strokes & Cosine & First & First 5 & Strokes \\
\midrule
all vectors (base) & 1 & 0.115 & 54.9\% & 73.3\% & 21.1 & 0.102 & 28.8\% & 51.2\% & 59.5 \\
 & 2 & 0.130 & 75.4\% & 89.6\% & 14.8 & 0.122 & 43.0\% & 70.0\% & 50.2 \\
all vectors, second seed & 1 & 0.118 & 57.5\% & 76.5\% & 23.3 & 0.105 & 31.2\% & 52.8\% & 59.7 \\
 & 2 & 0.130 & 75.1\% & 90.7\% & 13.0 & 0.122 & 43.7\% & 71.5\% & 44.7 \\
picture and label & 1 & 0.119 & 57.5\% & 77.4\% & 19.4 & 0.072 & 6.8\% & 16.8\% & 74.6 \\
 & 2 & 0.133 & 77.5\% & 91.7\% & 15.1 & 0.082 & 9.5\% & 24.0\% & 64.5 \\
picture only & 1 & 0.060 & 4.5\% & 11.6\% & 143.8 & 0.067 & 5.0\% & 12.2\% & 149.7 \\
 & 2 & 0.063 & 6.1\% & 14.2\% & 84.1 & 0.073 & 4.3\% & 17.2\% & 90.6 \\
picture only, shifted & 1 & 0.081 & 18.8\% & 38.0\% & 60.9 & 0.087 & 11.0\% & 26.7\% & 114.5 \\
 & 2 & 0.084 & 22.3\% & 43.8\% & 26.3 & 0.097 & 17.3\% & 36.8\% & 54.7 \\
no label dropout & 1 & 0.119 & 57.7\% & 77.4\% & 14.5 & 0.110 & 32.7\% & 58.7\% & 44.8 \\
no augmentation & 1 & 0.078 & 14.3\% & 28.8\% & 71.2 & 0.093 & 17.3\% & 37.5\% & 75.8 \\
 & 2 & 0.061 & 3.9\% & 11.7\% & 3.9 & 0.065 & 6.2\% & 17.7\% & 3.4 \\
\bottomrule
\end{tabular}
\end{small}
\end{center}
\end{table}

\begin{table}[H]
\caption{Where the vectors lie. Pictures of sketches sit in a narrow cone, nearly at right angles to text. SigLIP~2 vectors of 1,000 held-out drawings and their text. One picture's cosine to another picture is 0.726, and the mean picture's cosine to the mean label is 0.044. The kinds of caption are described in Appendix \ref{app:pipeline}. Told is the nats per pen move of an earlier model of the base recipe when told that vector alone (0.6121 told nothing, 0.5972 told the picture).}
\label{tab:geometry}
\vskip 0.1in
\begin{center}
\begin{small}
\begin{tabular}{lcccc}
\toprule
Text & Own picture & Another's & Own nearest & Told \\
\midrule
label & 0.114 & 0.033 & 28.6\% & 0.6036 \\
frame & 0.193 & 0.050 & 56.5\% & 0.6001 \\
phrase & 0.134 & 0.000 & 43.8\% & 0.6017 \\
short & 0.147 & 0.002 & 50.9\% & 0.6009 \\
dense & 0.159 & 0.033 & 47.2\% & 0.6001 \\
sentence & 0.169 & 0.034 & 46.5\% & 0.6006 \\
\bottomrule
\end{tabular}
\end{small}
\end{center}
\end{table}

\begin{figure}[H]
    \centering \includegraphics[width=\textwidth]{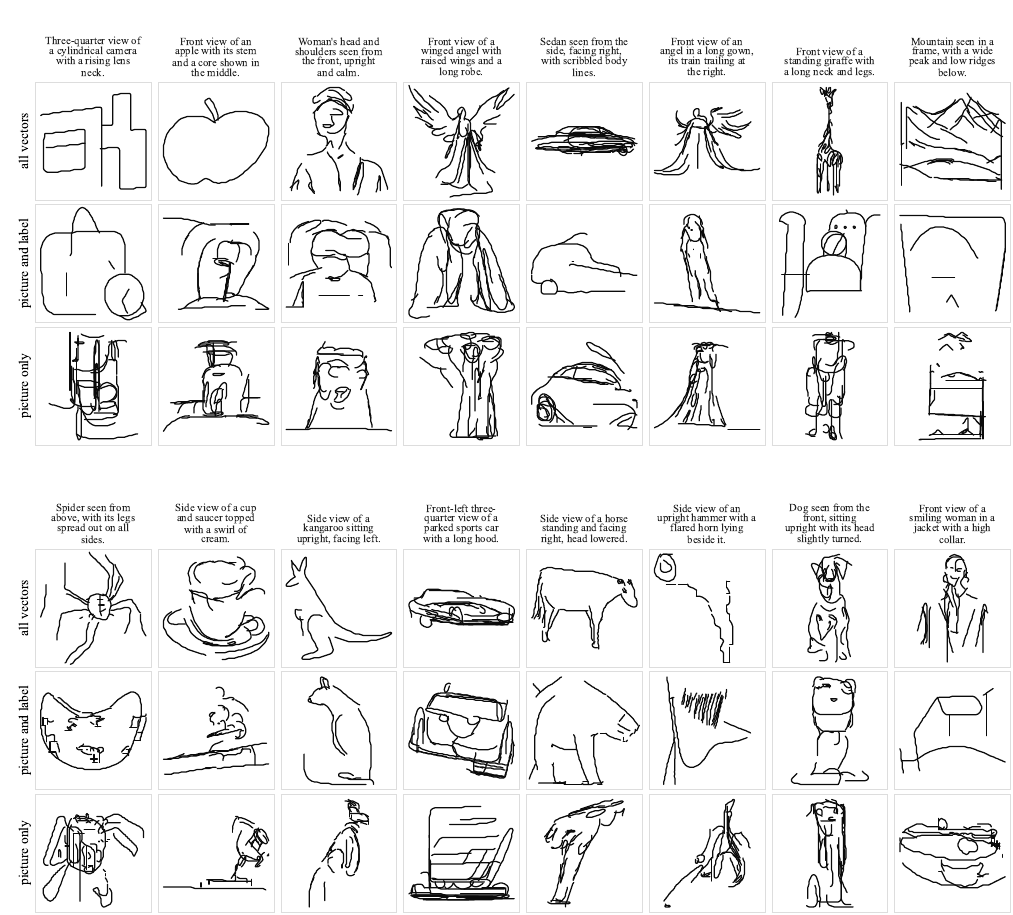}
    \caption{Drawing from caption phrases. The first sixteen held-out caption phrases, each drawn once by the three models of the caption study at guidance 2 and temperature $0.6$, with nothing picked or redrawn. Only the model trained on all vectors follows the phrase.}
    \label{fig:captions}
\end{figure}

\section{More samples}

\begin{figure}[H]
    \centering \includegraphics[width=.6\textwidth]{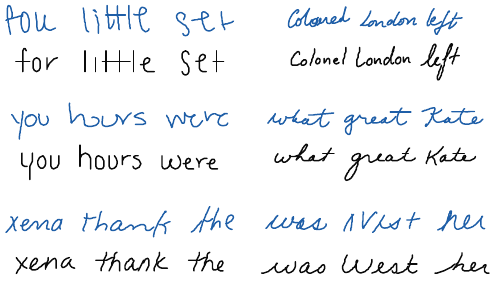}
    \caption{Taking up an unseen hand. Three words by each of six writers held out of training (blue) and, under each, the same words written by the model after it has read one to three \emph{other} words by that writer (black).}
    \label{fig:transfer}
\end{figure}

\begin{figure}[H]
    \includegraphics[width=\textwidth]{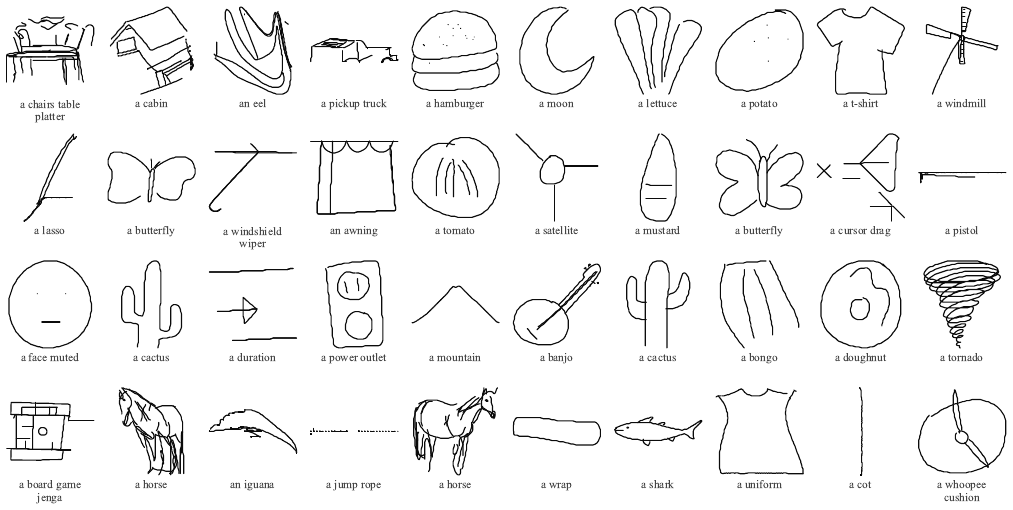}
    \caption{Random samples from the drawing model. The labels of 40 held-out drawings picked at random from seven sources, one draw of each at temperature $0.4$ and guidance 2, with nothing picked or redrawn.}
    \label{fig:random}
\end{figure}

\begin{figure}[H]
    \includegraphics[width=\textwidth]{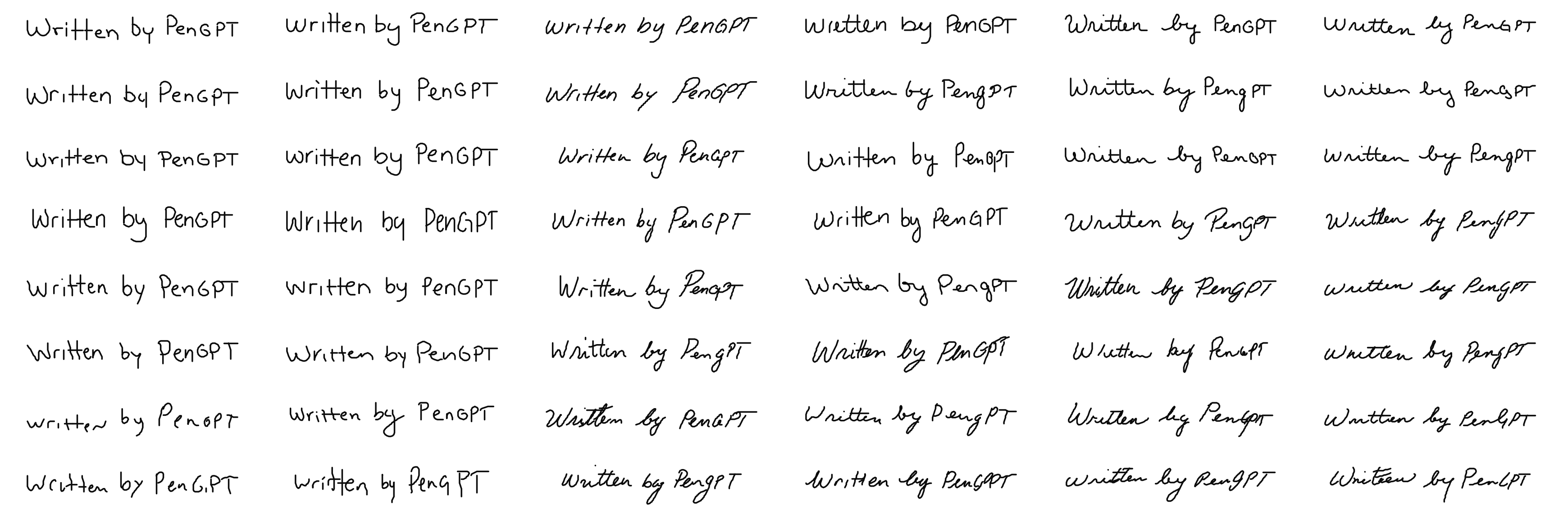}
    \vspace{-1.3\baselineskip}
    \caption{Styles the model knows. One phrase in 48 of the model's 1,192 hands, print to cursive across and neat to messy down.}
    \label{fig:styles}
    \vspace{-0.7\baselineskip}
\end{figure}

\begin{figure}[H]
    \centering \includegraphics[width=.8\textwidth]{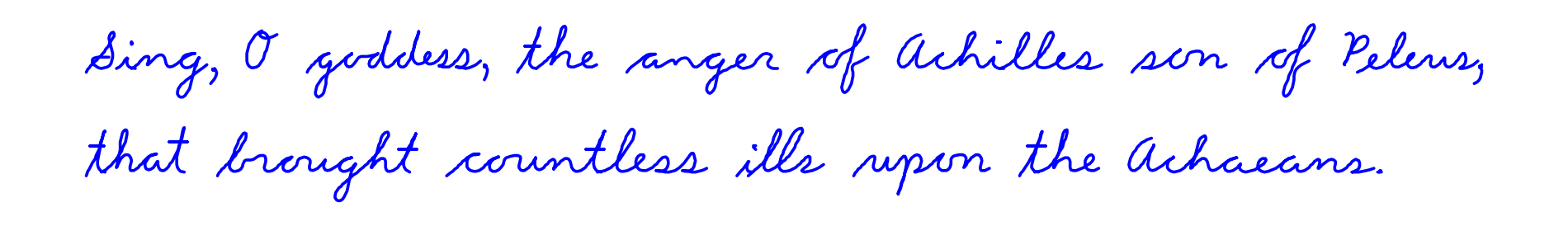}
    \caption{A long passage. The opening of the Iliad, written two words per call with each call continuing from the last word written, so that the hand stays steady over the passage.}
    \label{fig:iliad}
\end{figure}

\begin{figure}[H]
    \centering \includegraphics[width=.8\textwidth]{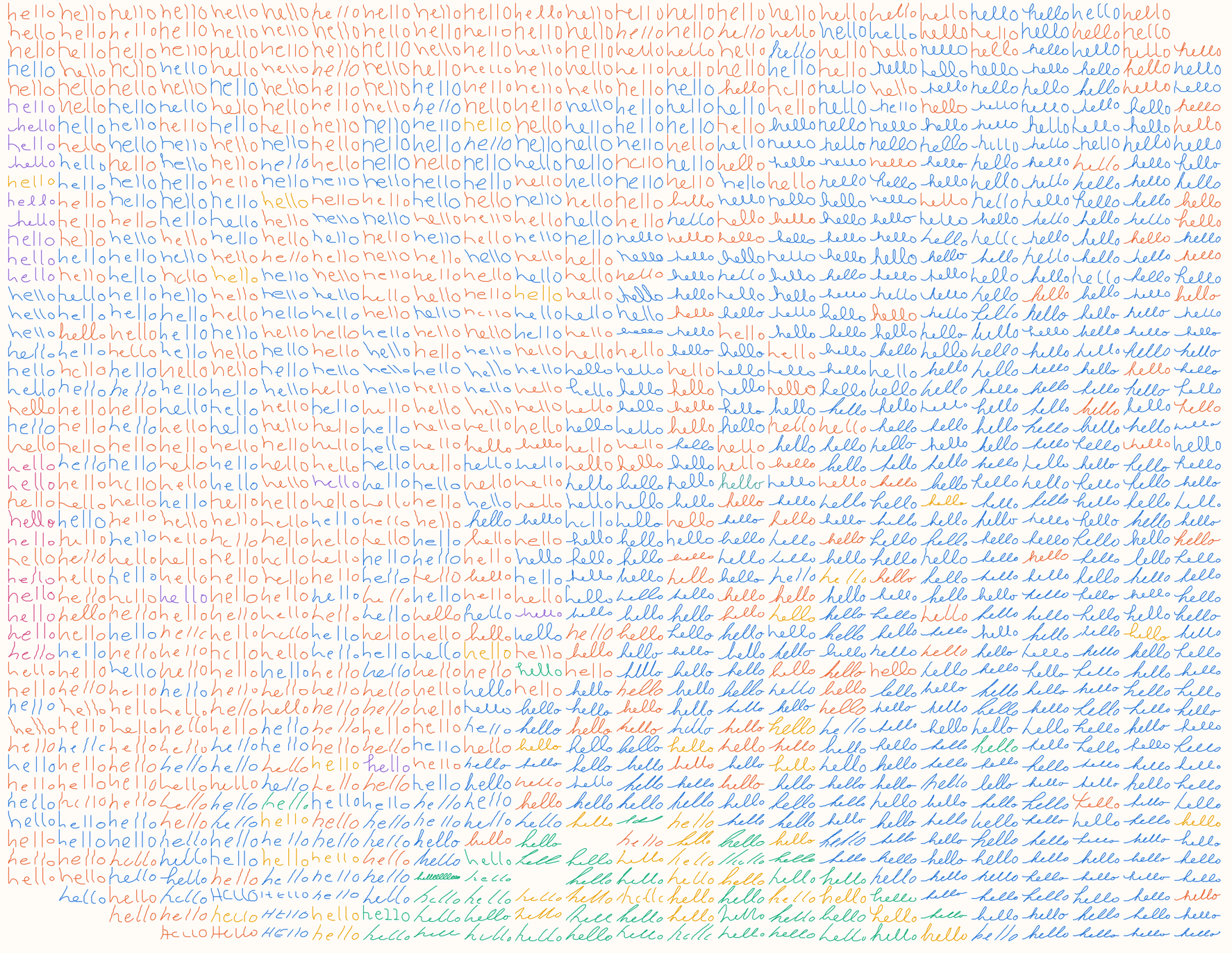}
    \caption{A map of hands. Every writer's learned vector, placed by UMAP, moved to the nearest free cell of a grid and written as that writer's ``hello'': print on the left, cursive on the right. Colour is the source a writer's pages came from.}
    \label{fig:writers}
\end{figure}

\end{document}